\documentclass[sigconf, nonacm]{acmart}
\renewcommand\footnotetextcopyrightpermission[1]{}
\usepackage{romannum}
\usepackage{float}
\usepackage{subcaption}  
\usepackage{tikz}
\usepackage{pgfplots}
\pgfplotsset{compat=1.18}

\begin{document}

\title{HumanoidVLM: Vision–Language–Guided Impedance Control for Contact-Rich Humanoid Manipulation}



\author{Yara Mahmoud}
\affiliation{%
  \institution{Skolkovo Institute of Science and Technology}
  \city{Moscow}
  \country{Russia}}
\email{yara.mahmoud@skoltech.ru}

\author{Yasheerah Yaqoot}
\affiliation{%
  \institution{Skolkovo Institute of Science and Technology}
  \city{Moscow}
  \country{Russia}}
\email{yasheerah.yaqoot@skoltech.ru}

\author{Miguel Altamirano Cabrera}
\affiliation{%
  \institution{Skolkovo Institute of Science and Technology}
  \city{Moscow}
  \country{Russia}}
\email{m.altamirano@skoltech.ru}

\author{Dzmitry Tsetserukou}
\affiliation{%
  \institution{Skolkovo Institute of Science and Technology}
  \city{Moscow}
  \country{Russia}}
\email{d.tsetserukou@skoltech.ru}


\begin{teaserfigure}
  \includegraphics[width=\textwidth]{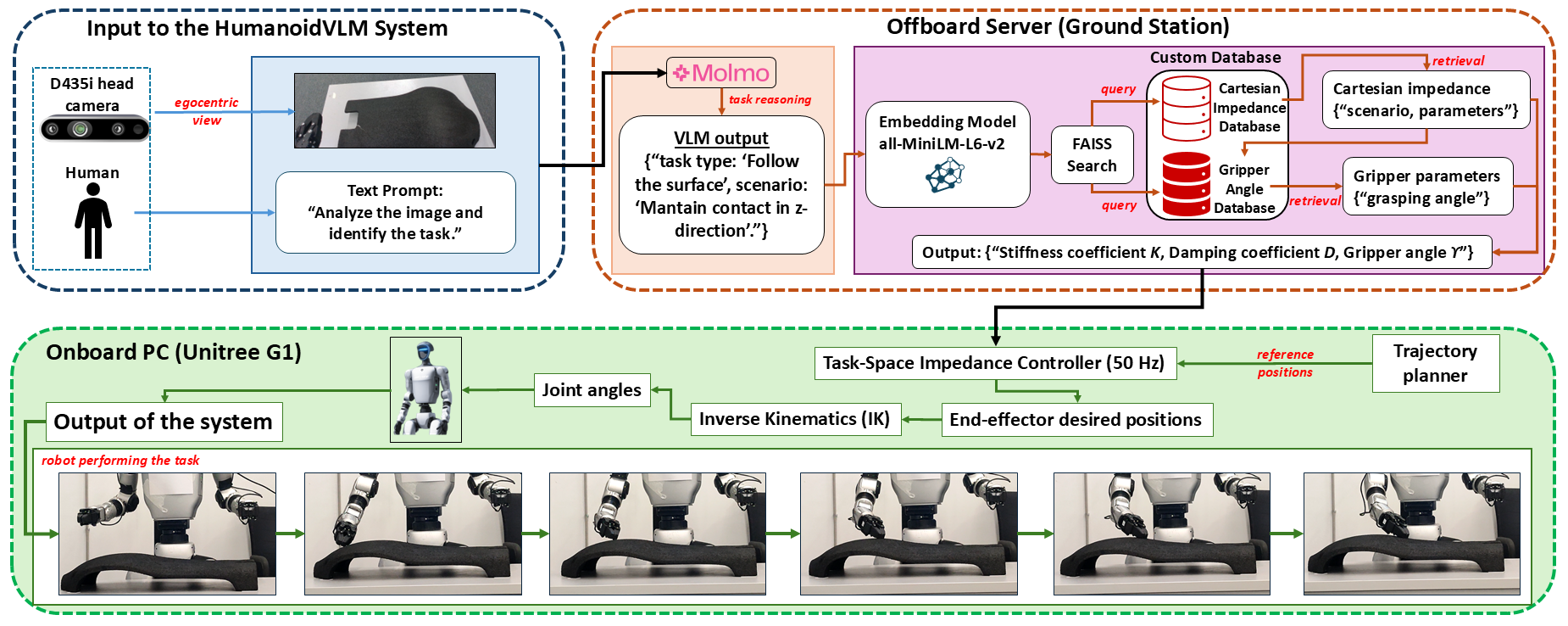}
  \caption{Overall architecture of the proposed HumanoidVLM framework.
  }
  \Description{A schematic diagram showing the full HumanoidVLM pipeline. On the left, a human and an Intel RealSense D435i head-mounted camera produce an egocentric RGB image. In the top center, this image flows into a blue box labeled “Input to the HumanoidVLM System,” which contains an example camera frame. The image is routed to the Molmo vision–language model, shown in a pink box labeled “task reasoning,” which outputs a structured task description. The output is then passed to a purple box labeled “Offboard Server (Ground Station),” which includes an embedding model, a FAISS search module, and two stacked database icons representing the cartesian impedance database and the gripper-angle database. Arrows indicate retrieval of stiffness and damping parameters and a gripper angle. These values are transmitted to a green box labeled “Onboard PC (Unitree G1),” which contains blocks for the task-space impedance controller, inverse kinematics module, and trajectory planner. At the bottom, a sequence of six photos shows the robot performing the task, moving its right arm along an object on a table.
}
  \label{fig:sys-arch}
  \vspace{-0.1cm}
\end{teaserfigure}

\begin{abstract}
Humanoid robots must adapt their contact behavior to diverse objects and tasks, yet most controllers rely on fixed, hand-tuned impedance gains and gripper settings. This paper introduces HumanoidVLM, a vision–language driven retrieval framework that enables the Unitree G1 humanoid to select task-appropriate cartesian impedance parameters and gripper configurations directly from an ego-centric RGB image. The system couples a vision–language model for semantic task inference with a FAISS-based Retrieval-Augmented Generation (RAG) module that retrieves experimentally validated stiffness–damping pairs and object-specific grasp angles from two custom databases, and executes them through a task-space impedance controller for compliant manipulation. We evaluate HumanoidVLM on 14 visual scenarios and achieve a retrieval accuracy of 93\%. Real-world experiments show stable interaction dynamics, with $z$-axis tracking errors typically within \(1\text{--}3.5\) cm and virtual forces consistent with task-dependent impedance settings. These results demonstrate the feasibility of linking semantic perception with retrieval-based control as an interpretable path toward adaptive humanoid manipulation.
\vspace{-0.2cm}
\end{abstract}


\vspace{-0.9cm}

\maketitle
\vspace{-0.5cm}
\section{Introduction}
Humanoid robots are increasingly expected to operate in unstructured human environments, where tasks such as placing objects, applying force on surfaces, grasping diverse items, or manipulating tools require a combination of semantic understanding and compliant physical interaction \cite{bicchi2000roboticHands}. Conventional control pipelines for such systems typically rely on fixed, hand-tuned impedance gains \cite{hogan1985_impedance_i} and manually specified gripper configurations, which restrict the robot’s ability to adapt to changes in scene geometry, object properties, or task intent.  

In parallel, recent advances in Vision--Language Models (VLMs) have demonstrated impressive capabilities in open-world perception, contextual reasoning, and grounded language understanding \cite{alayrac2022flamingo}, making them promising candidates for high-level task inference in robotics \cite{brohan2023rt1roboticstransformerrealworld}. However, despite their strong semantic understanding, these models remain largely disconnected from the low-level control parameters that determine how a robot physically interacts with its environment. As a result, current humanoid systems lack the capacity to autonomously translate visual semantics into actionable, task-appropriate interaction behaviors—a limitation that is especially pronounced in contact-rich manipulation tasks, where selecting correct stiffness, damping, and grasping configurations is essential for safe and stable execution.

To bridge this gap, we introduce HumanoidVLM, a retrieval-augmented vision--language framework that connects high-level visual reasoning with low-level manipulation parameters for the Unitree~G1 humanoid robot. Given a single egocentric image, HumanoidVLM infers the ongoing manipulation task through structured visual queries and retrieves the corresponding cartesian impe-dance parameters and object-specific gripper angle from custom, experimentally validated databases. This approach enables the robot to autonomously determine task-appropriate compliance settings for a given scene.
\vspace{-0.3cm}

\section{Related Works}
\textbf{Foundation models and semantic manipulation.}Recent advances in foundation models have accelerated progress in semantic manipulation and scene-aware task execution. OmniVIC ~\cite{omnivic2025_vlm_vic} and ImpedanceGPT~\cite{batool2025impedancegpt} are most closely related to our approach, combining a vision--language model with a variable impedance controller through retrieval. However, OmniVIC focuses on industrial arms and does not address humanoid embodiment, shared workspaces, or HRI-oriented safety. Vision--language--action systems such as PG-VLM~\cite{gao2024_pgvlm}, SayCan~\cite{ahn2023_saycan}, RT-2~\cite{brohan2023_rt2},Bi-VLA ~\cite{gbagbe2024bivla}, and SwarmVLM ~\cite{zafar2025swarmvlm} show that multimodal models can generalize manipulation behavior, but these frameworks operate at the policy or pose level and rely on fixed impedance gains. Broader surveys on foundation models for manipulation and robot intelligence~\cite{li2025_fm_manip_survey}, \cite{jeong2024_llm_robot_intel} highlight the role of semantic reasoning but do not consider continuous impedance modulation in humanoid settings. Work on foundation models for collaborative assembly~\cite{ji2024_fm_hrc} emphasizes contextual reasoning in shared autonomy, yet compliant control remains largely unaddressed.

\textbf{Impedance and variable impedance control.} Classical impe-dance control, introduced by Hogan~\cite{hogan1985_impedance_i}, provides a structured formulation for regulating robot--environment interaction through stiffness and damping shaping. Surveys by Abu-Dakka and Saveriano~\cite{abudakka2020_vic_review} document advances in variable impedance strategies, including learning-based adaptation and physical interaction applications. In human--robot collaboration, Ajoudani~\emph{et~al.}~\cite{ajoudani2018_hrc} stress the importance of compliance modulation for safety and ergonomics, following earlier studies on tactile and impedance-based interaction in humanoid systems ~\cite{tsetserukou2008tactile}. Building on this foundation, our work links semantic scene understanding to impedance and gripper selection through retrieval-augmented reasoning, enabling task-aware compliant control on a humanoid.

\vspace{-0.2cm}
\section{System Architecture}
The proposed system enables the Unitree G1 humanoid to adapt its end-effector impedance and grasping behavior using visual scene understanding and retrieval-based reasoning. An ego-centric RGB image from the robot’s head camera is processed by a vision--language model (VLM), which infers the high-level manipulation task through structured visual queries. The resulting semantic task label is embedded and passed to a Retrieval-Augmented Generation (RAG) module that performs FAISS-based similarity search over two custom databases: (i) a cartesian impedance database storing task-specific stiffness and damping coefficients for the end-effector, and (ii) a gripper-angle database specifying the optimal grasp configuration for the object category.

The retrieved control parameters ($K$, $D$, and gripper angle $\gamma$) are transmitted to the onboard G1 computer, where a task-space cartesian impedance controller generates compliant end-effector trajectories. These desired virtual poses are then converted into joint targets through inverse kinematics and executed using the robot’s built-in position controllers. Fig.~\ref{fig:sys-arch} provides an overview of the perception, reasoning, and control pipeline.

\vspace{-0.2cm}
\subsection{VLM--RAG System}
\label{vlm-rag}
The VLM--RAG system processes the ego-centric view using the Molmo-7B-O BnB 4-bit model \cite{molmo_quantized} to identify the task-relevant objects and scene context. The visual output is converted into a multimodal representation and embedded using the \texttt{all-MiniLM-L6-v2} sentence-transformer \cite{sentence-embedding}, enabling semantically meaningful retrieval within a shared vector space. Given an input image, the VLM determines the task through sequential yes/no queries. The RAG system then retrieves the corresponding impedance and gripper parameters, providing the controller with context-dependent gains for safe and adaptive manipulation.

\subsubsection{Custom Database for Environmental Scenarios}
The database contains nine manipulation tasks, each associated with experimentally validated cartesian impedance parameters and an optimal gripper configuration. Impedance parameters include stiffness coefficients $K = [K_x, K_y, K_z]$ and damping coefficients $D = [D_x, D_y, D_z]$, which determine how the end-effector regulates contact forces and compliance along the cartesian axes. Each entry also specifies a preferred gripper angle $\gamma_a$ for secure manipulation of the corresponding object type. 

The retrieval system relies on two JSON databases. The impedance database stores task-specific cartesian impedance parameters collected from real-world experiments on the G1 humanoid. Multiple settings were tested per task, and the most stable and compliant configuration was recorded as the canonical entry. The gripper database contains experimentally determined optimal closing angles for manipulating rigid, soft, deformable, or fragile objects. These angles were obtained through repeated trials and represent secure yet compliant grasp strategies.

Together, these two databases map a semantic task label to its corresponding control variables, namely the cartesian impedance parameters and the gripper configuration, forming the knowledge base of the RAG module.
\vspace{-0.35cm}
\subsection{Robot Control}

\subsubsection{End-effector Impedance Control}
The Unitree G1 does not provide wrist force--torque sensing, and its two-finger grippers lack reaction wrench feedback. To enable compliant interaction without force sensors, we adopt a homogeneous cartesian mass--spring--damper model in task space. Each end-effector is regulated as a 6-DoF point following a desired pose trajectory, which acts as a virtual reference for the impedance dynamics.

The translational impedance parameters \( \mathbf{K} \) and \( \mathbf{D} \) are selected by the VLM--RAG system, while the virtual mass \( \mathbf{M} \) and rotational impedance remain fixed for stability. For each arm \( a \in \{L, R\} \), let \( \mathbf{x}_a \in \mathbb{R}^3 \) and
\( \mathbf{R}_a \in SO(3) \) denote the current end-effector position and
orientation, and let \( \mathbf{x}_{a,\mathrm{ref}}, \mathbf{R}_{a,\mathrm{ref}}
\) denote the desired reference pose. We define diagonal virtual mass, damping, and stiffness matrices:
\begin{equation}
\begin{aligned}
\mathbf{M}_a &= \mathrm{diag}(M_{a,x}, M_{a,y}, M_{a,z}), \\
\mathbf{D}_a &= \mathrm{diag}(D_{a,x}, D_{a,y}, D_{a,z}),\\
\mathbf{K}_a &= \mathrm{diag}(K_{a,x}, K_{a,y}, K_{a,z}).
\end{aligned}
\end{equation}
For each arm $a$, the translational error is defined as in~(2), the impedance dynamics follow the second order ODE~(3), and the resulting virtual force is given by~(4).
\begin{equation}
\mathbf{e}_a = \mathbf{x}_{a,\mathrm{ref}} - \mathbf{x}_a, \qquad
\dot{\mathbf{e}}_a = \dot{\mathbf{x}}_{a,\mathrm{ref}} - \dot{\mathbf{x}}_a
\end{equation}
\begin{equation}
\mathbf{M}_a \ddot{\mathbf{e}}_a + \mathbf{D}_a \dot{\mathbf{e}}_a + \mathbf{K}_a \mathbf{e}_a = \mathbf{0}
\end{equation}
\begin{equation}
\mathbf{F}_a^{\mathrm{virt}} = \mathbf{K}_a \mathbf{e}_a + \mathbf{D}_a \dot{\mathbf{e}}_a
\end{equation}



captures the interaction response, acting as a quantitative proxy for physical contact forces. If no disturbance exists, both pose error and the corresponding virtual forces converge to zero.

\subsubsection{Gripper Configuration}
Each hand uses a single-DoF gripper. The controller receives a discrete action: $\gamma_a \in \{\text{open}, \text{close}\}$, where each action corresponds to a predefined joint angle target retrieved from the database. Gripper angles are retrieved from the VLM--RAG database based on the inferred task and object type, complementing the impedance parameters selected for the scenario.

\vspace{-0.2cm}
\section{System Evaluation}
Experiments were conducted on the Unitree G1 humanoid robot equipped with two 1-DoF grippers and an Intel RealSense RGB-D camera mounted in the head for ego-centric perception. The task-space impedance controller ran on the on-board PC at 50\,Hz, while the VLM--RAG pipeline was executed on an external workstation with an RTX~4090 GPU and Intel i9-13900K CPU to ensure real-time inference.
\begin{figure}[t]
    \centering
    \includegraphics[width=0.8\linewidth]{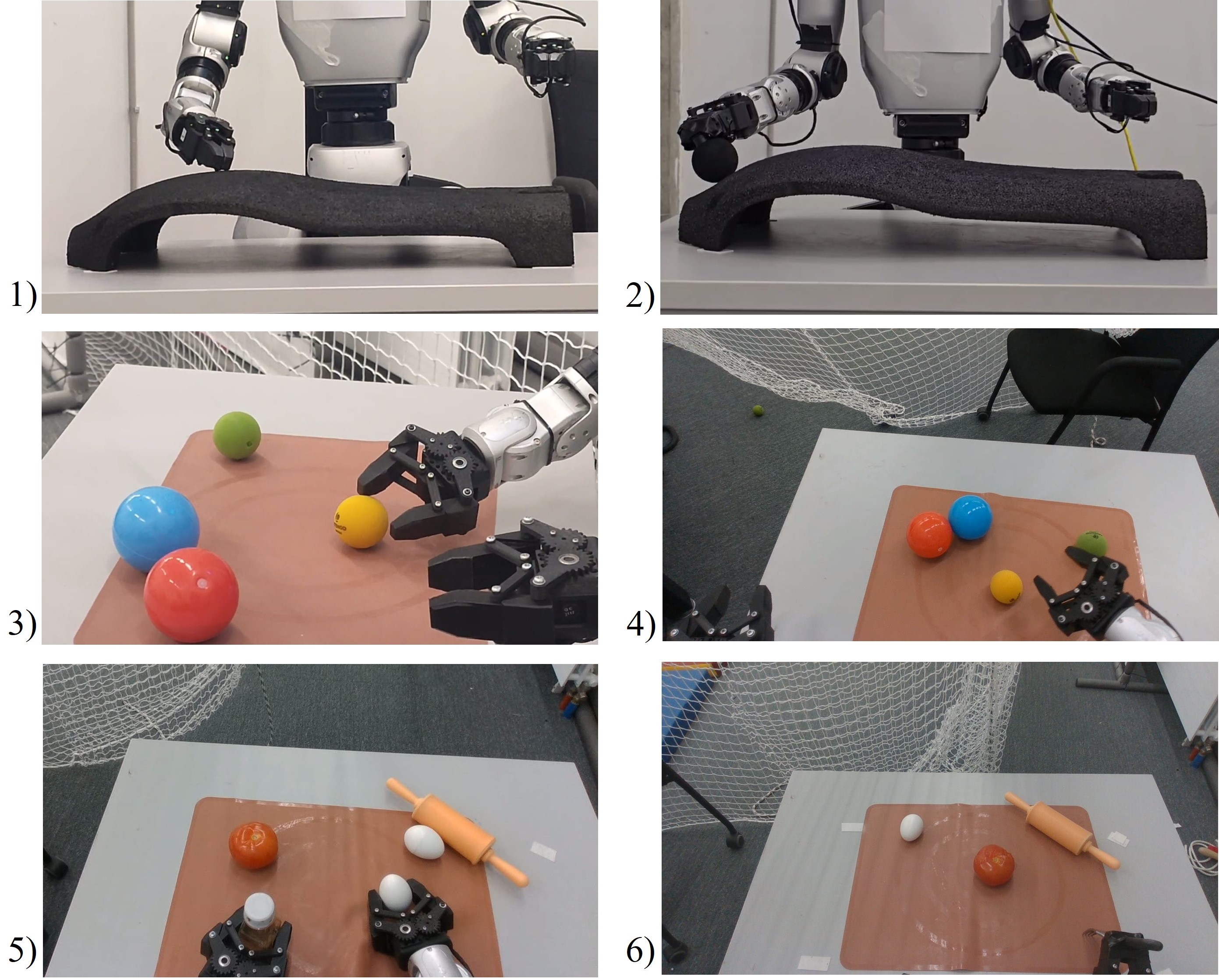}
    \caption{Images from successful trials of the representative manipulation scenarios.}
    \label{fig:scenario}
    \Description{A collage of six separate images illustrating the manipulation tasks used in the experiments.  Image 1: A frontal view of the G1 humanoid’s upper body, with the right hand gliding along a curved tabletop object for surface following. Image 2: The right arm holding a massage ball and pressing it against an object on a table. Image 3: A side view of balls scattered on a table while the robot’s right hand reaches toward a ball to grasp it. Image 4: An egocentric camera view showing the same grasping scene from the robot’s perspective. Image 5: An egocentric view of a bimanual placement task, with the left hand holding a small bottle and the right hand holding an egg. Image 6: A top-down table view from the robot’s head camera showing the right hand holding a fork and poking a tomato.}
\vspace{-0.65cm}
\end{figure}
\vspace{-0.4cm}
\subsection{Experimental Setup}
Evaluation focused on tabletop tasks dominated by normal-direction interaction along the $z$-axis. Five representative task groups were considered: (1) Following an irregular surface by maintaining smooth right-hand contact, (2) Applying controlled $z$-direction forces while holding a massage ball, (3) Bimanual placement of two objects: a sauce bottle (left) and an egg (right), (4) Tool interaction using a fork to poke a fruit or vegetable, and (5) Grasping and lifting various tabletop objects.

Each scenario was executed with multiple impedance and gripper settings to empirically determine the optimal parameters stored in the database. Fig.~\ref{fig:scenario} shows examples from successful trials.

\subsubsection{VLM--RAG Retrieval Evaluation}
\label{subsec:vlm_rag_eval}
The evaluation tests the discrete retrieval correctness of impedance and gripper parameters from a single ego-centric image.
A retrieval is considered correct for evaluation purposes when the system: (1) classifies the task correctly via sequential visual yes/no queries, (2) retrieves the correct impedance entry based on the VLM output, and (3) retrieves the correct gripper configuration conditioned on both the VLM task label and the impedance scenario.

\subsubsection{Retrieval Process}
To assess semantic generalization across tasks, 14 ego-centric test images were collected. These depict the nine task types but vary in camera viewpoint, object placement, and arm pose, ensuring robustness evaluation rather than memorization. For each image, the VLM infers the task through hierarchical visual queries. The resulting task label is used as a FAISS query to retrieve the impedance scenario. The scenario description is then concatenated with the VLM label to form a second query for retrieving the appropriate gripper configuration. This two-stage retrieval disambiguates visually similar tasks differing in object type or interaction mode.

\subsubsection{Impedance Control Evaluation}
\label{sec:imp_eval}
To confirm that retrieved parameters yield stable and bounded interaction behavior, we analyze $z$-axis tracking errors and virtual force magnitudes during execution. Single-object tasks use the right arm; dual-object placement uses both arms. The evaluation checks whether the selected $K_z$ and $D_z$ values achieve the expected compliance and whether position errors and virtual forces remain bounded and task-appropriate.
\vspace{-0.2cm}

\subsection{Results}

\subsubsection{VLM--RAG System}
\begin{figure}[h]
    \centering
    \includegraphics[width=0.8\linewidth]{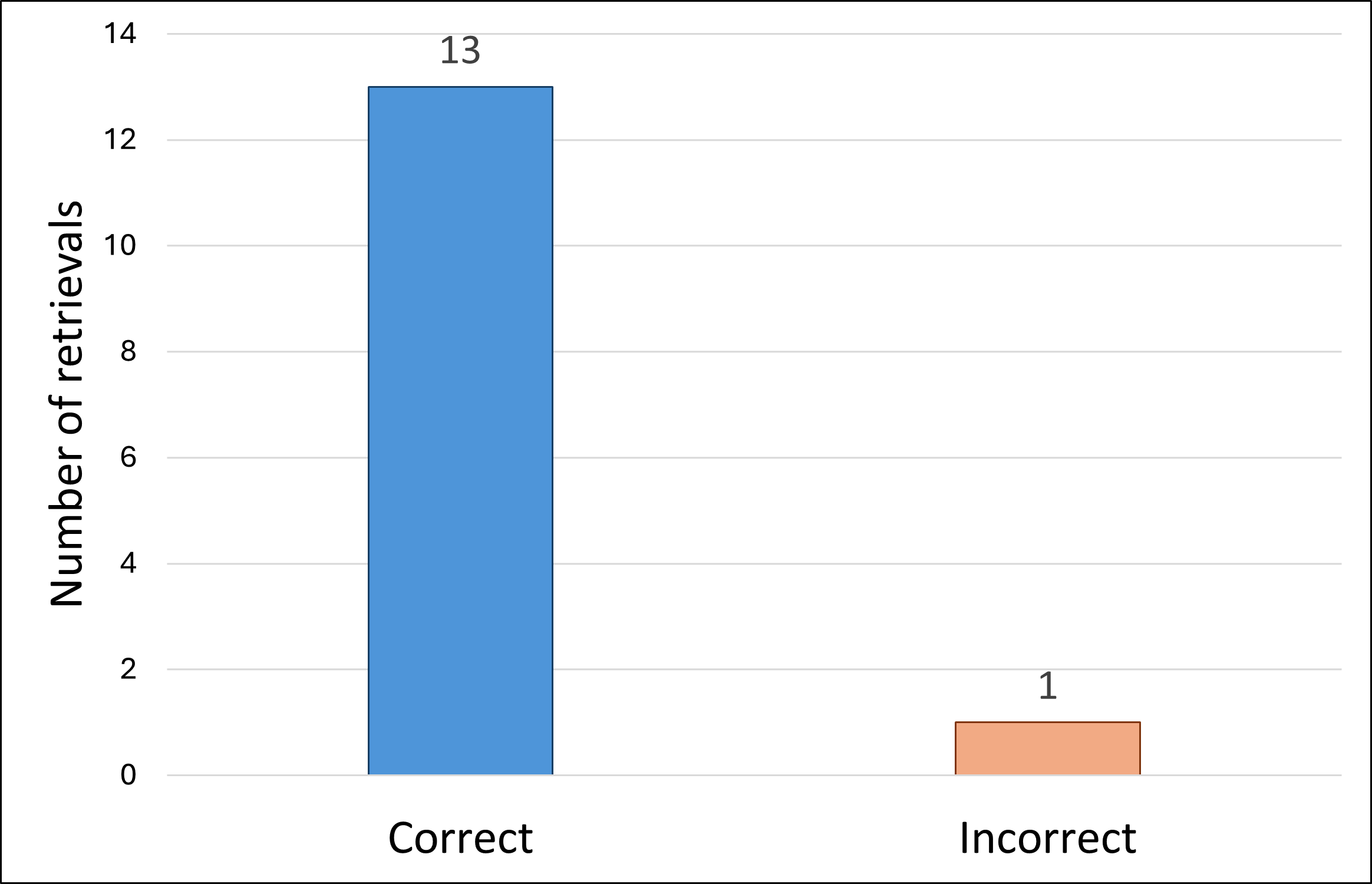}
    \caption{Retrieval accuracy of the VLM--RAG system across 14 scenarios.}
    \Description{A bar chart with two vertical bars. The left bar, labeled “Correct,” has height 13. The right bar, labeled “Incorrect,” has height 1. The horizontal axis contains the two categories, and the vertical axis shows the number of retrievals.}
    \label{fig:vlm-rag-result}
    \vspace{-0.2cm}
\end{figure}

The VLM--RAG pipeline was evaluated on 14 ego-centric test images representing variations of the nine tasks stored in the database (Sec.~\ref{subsec:vlm_rag_eval}). The system retrieved the correct impedance and gripper parameters in 13 out of 14 cases, resulting in an accuracy of 93\% as can be seen in Fig.~\ref{fig:vlm-rag-result}. The single failure occurred when the primary object was partially occluded, highlighting a limitation of the current vision-based task inference. 

\subsubsection{Control Evaluation Results}
\label{sec:ctrl_results}

Table~\ref{tab:impedance_eval} summarizes the retrieved parameters and corresponding performance metrics. For each scenario, we report the selected normal-direction gains $K_z, D_z$, the mean and maximum absolute position error in $z$, and the maximum virtual normal force magnitude. Across all scenarios, $z$-axis tracking errors remained small, and the computed virtual force magnitudes scaled consistently with the selected impedance gains. This confirms that the retrieved parameters produce stable and task-appropriate interaction behavior.
\begin{table}[h!]
\caption{Task-space impedance evaluation for representative tabletop tasks.}
\label{tab:impedance_eval}
\small
\setlength{\tabcolsep}{4pt}
\renewcommand{\arraystretch}{1.05}
\begin{tabular}{l c c c c c}
\toprule
Task &
$K_z$ &
$D_z$ &
$\overline{|e_z|}$ &
$\max |e_z|$ &
$\max |F^{\mathrm{virt}}_z|$ \\
 &
[N/m] &
[Ns/m] &
[m] &
[m] &
[arb.] \\
\midrule
Follow surface (R)   & 3.0 & 2.0 & 0.016 & 0.024 & 0.103 \\
apply pressure (R)   & 5.0 & 3.0 & 0.017 & 0.035 & 0.334 \\
Dual placement (R)        & 2.0 & 1.0 & 0.013 & 0.034 & 0.089 \\
Dual placement (L)     & 6.0 & 1.5 & 0.009 & 0.013 & 0.176 \\
Tool interaction (R)  & 2.0 & 1.5 & 0.006 & 0.016 & 0.183 \\
Grasp from table (R)           & 4.0 & 1.5 & 0.013 & 0.022 & 0.153 \\
\bottomrule
\Description{ A table listing six tasks (surface following, massage, dual placement for egg and bottle, tool poking, and grasping). For each, the table reports the arm used, stiffness K_z, damping D_z, mean and maximum absolute position error in z, and maximum virtual force magnitude.}
\end{tabular}
\vspace{-0.7cm}
\end{table}

\begin{figure}[t]
    \raggedleft
    \includegraphics[width=0.43\textwidth]{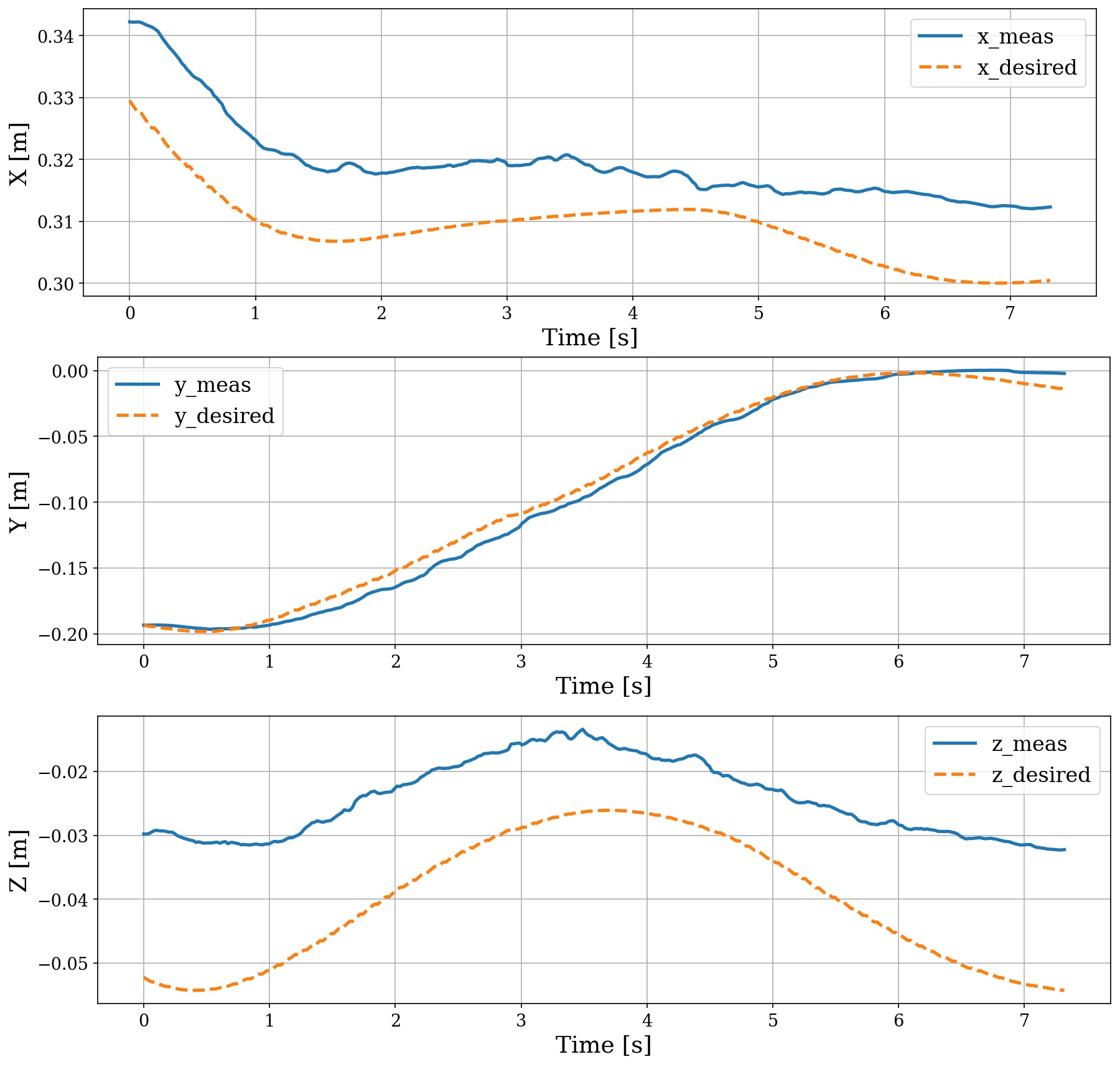}
    \caption{Translational errors in the surface-following task, showing desired and measured end-effector positions using forward kinematics.}
    \Description{A set of three line plots showing end-effector position tracking over time for the surface-following task.  Top plot: measured and desired X-position curves.  Middle plot: measured and desired Y-position curves.  Bottom plot: measured and desired Z-position curves.  Each plot contains a solid blue line for measured motion and an orange dashed line for the reference trajectory, with time in seconds on the horizontal axis.}
    \label{fig:pos_error}
    \vspace{-0.35cm}
\end{figure}

\textbf{Surface following} Soft stiffness enabled compliant tracking of curved geometry, with low virtual forces and small errors. As seen in Fig.~\ref{fig:pos_error}, the robot tracks the curved object with compliant contact. Fig.~\ref{fig:pos_error} shows that $z$ error follow the surface profile, while $x$ deviations stem from orientation adjustments of the arm. \textbf{Massage and pressure application:} Higher stiffness and damping resulted in larger virtual forces while maintaining similar tracking accuracy, demonstrating effective modulation of contact intensity. \textbf{Dual-object placement:} Asymmetric stiffness values reflected object fragility. The egg-handling arm used soft settings, while the bottle-handling arm used stiffer gains, producing correspondingly higher force responses. \textbf{Tool interaction:} Moderate stiffness and damping supported controlled poking motions with low tracking error and transient force peaks appropriate for brief contact events. \textbf{Grasp-and-lift:} Intermediate stiffness ensured stable lifting without excessive pressure, with tracking errors remaining below 2.5\,cm.

The controller behaved consistently, with low stiffness enabling compliant contact, high stiffness supporting forceful interaction, and asymmetric bimanual settings matching object-specific needs. The clear link between impedance parameters and virtual force responses indicates that these virtual forces may serve as effective proxies for real contact forces.

\section{Conclusion and Future Work}
This paper presented HumanoidVLM, a vision–language–driven retrieval framework that links semantic scene understanding with task-specific cartesian impedance and gripper control for the Unitree G1 humanoid. By combining VLM-based task inference with a FAISS-based RAG module, the system autonomously selects stiffness, damping, and grasp configurations directly from an ego-centric image using two small human-validated databases containing nine manipulation tasks and nine object-specific grasp entries. Experiments across manipulation tasks showed that the framework retrieves correct parameters in 93\% of the evaluated scenarios and enables stable, compliant execution in real-world trials.

Given the limited evaluation scale and task diversity, the current results should be interpreted as a proof of feasibility rather than a comprehensive robustness assessment. In future work, we aim to incorporate task-dependent \emph{rotational} impedance to enable orientation-sensitive behaviors. Second, substituting the discrete database with a learned continuous mapping could allow interpolation across unseen tasks and object types. Finally, integrating force or visuotactile feedback would allow closed-loop impedance adaptation, improving safety and robustness.

\vspace{-0.2cm}
\section*{Acknowledgements} 
Research reported in this publication was financially supported by the RSF grant No. 24-41-02039.

\balance 

\begin{thebibliography}{19}


\ifx \showCODEN    \undefined \def \showCODEN     #1{\unskip}     \fi
\ifx \showISBNx    \undefined \def \showISBNx     #1{\unskip}     \fi
\ifx \showISBNxiii \undefined \def \showISBNxiii  #1{\unskip}     \fi
\ifx \showISSN     \undefined \def \showISSN      #1{\unskip}     \fi
\ifx \showLCCN     \undefined \def \showLCCN      #1{\unskip}     \fi
\ifx \shownote     \undefined \def \shownote      #1{#1}          \fi
\ifx \showarticletitle \undefined \def \showarticletitle #1{#1}   \fi
\ifx \showURL      \undefined \def \showURL       {\relax}        \fi
\providecommand\bibfield[2]{#2}
\providecommand\bibinfo[2]{#2}
\providecommand\natexlab[1]{#1}
\providecommand\showeprint[2][]{arXiv:#2}

\bibitem[mol(2025)]%
        {molmo_quantized}
 \bibinfo{year}{2025}\natexlab{}.
\newblock \bibinfo{booktitle}{\emph{{Molmo-7B-O} {BnB} 4bit quantized 7GB}}.
\newblock
\urldef\tempurl%
\url{https://huggingface.co/cyan2k/molmo-7B-O-bnb-4bit}
\showURL{%
\tempurl}


\bibitem[sen(2025)]%
        {sentence-embedding}
 \bibinfo{year}{2025}\natexlab{}.
\newblock \bibinfo{booktitle}{\emph{Sentence Transformer: all-{MiniLM}-{L6}-v2}}.
\newblock
\urldef\tempurl%
\url{https://huggingface.co/sentence-transformers/all-MiniLM-L6-v2}
\showURL{%
\tempurl}


\bibitem[Abu-Dakka and Saveriano(2020)]%
        {abudakka2020_vic_review}
\bibfield{author}{\bibinfo{person}{Fares~J. Abu-Dakka} {and} \bibinfo{person}{Matteo Saveriano}.} \bibinfo{year}{2020}\natexlab{}.
\newblock \showarticletitle{Variable Impedance Control and Learning—A Review}.
\newblock \bibinfo{journal}{\emph{Frontiers in Robotics and AI}}  \bibinfo{volume}{7} (\bibinfo{year}{2020}), \bibinfo{pages}{590681}.
\newblock
\href{https://doi.org/10.3389/frobt.2020.590681}{doi:\nolinkurl{10.3389/frobt.2020.590681}}


\bibitem[Ahn et~al\mbox{.}(2023)]%
        {ahn2023_saycan}
\bibfield{author}{\bibinfo{person}{Michael Ahn}, \bibinfo{person}{Anthony Brohan}, \bibinfo{person}{Noah Brown}, \bibinfo{person}{Yevgen Chebotar}, \bibinfo{person}{Omar Cortes}, \bibinfo{person}{Byron David}, \bibinfo{person}{Chelsea Finn}, \bibinfo{person}{Chuyuan Fu}, \bibinfo{person}{Keerthana Gopalakrishnan}, \bibinfo{person}{Karol Hausman}, {et~al\mbox{.}}} \bibinfo{year}{2023}\natexlab{}.
\newblock \showarticletitle{Do As I Can, Not As I Say: Grounding Language in Robotic Affordances}.
\newblock \bibinfo{journal}{\emph{Proc. of the Conf. on Robot Learning}}  \bibinfo{volume}{205} (\bibinfo{year}{2023}).
\newblock


\bibitem[Ajoudani et~al\mbox{.}(2018)]%
        {ajoudani2018_hrc}
\bibfield{author}{\bibinfo{person}{Arash Ajoudani}, \bibinfo{person}{Andrea~Maria Zanchettin}, \bibinfo{person}{Serena Ivaldi}, \bibinfo{person}{Alin Albu-Sch{\"a}ffer}, \bibinfo{person}{Kazuhiro Kosuge}, {and} \bibinfo{person}{Oussama Khatib}.} \bibinfo{year}{2018}\natexlab{}.
\newblock \showarticletitle{Progress and Prospects of the Human--Robot Collaboration}.
\newblock \bibinfo{journal}{\emph{Autonomous Robots}} \bibinfo{volume}{42}, \bibinfo{number}{5} (\bibinfo{year}{2018}), \bibinfo{pages}{957--975}.
\newblock
\href{https://doi.org/10.1007/s10514-017-9677-2}{doi:\nolinkurl{10.1007/s10514-017-9677-2}}


\bibitem[Alayrac et~al\mbox{.}(2022)]%
        {alayrac2022flamingo}
\bibfield{author}{\bibinfo{person}{Jean-Baptiste Alayrac}, \bibinfo{person}{Jeff Donahue}, \bibinfo{person}{Pauline Luc}, \bibinfo{person}{Antoine Miech}, \bibinfo{person}{Iain Barr}, \bibinfo{person}{Yana Hasson}, \bibinfo{person}{Karel Lenc}, \bibinfo{person}{Arthur Mensch}, \bibinfo{person}{Katherine Millican}, \bibinfo{person}{Malcolm Reynolds}, {et~al\mbox{.}}} \bibinfo{year}{2022}\natexlab{}.
\newblock \showarticletitle{Flamingo: a Visual Language Model for Few-Shot Learning}. In \bibinfo{booktitle}{\emph{Advances in Neural Information Processing Systems}}, \bibfield{editor}{\bibinfo{person}{S.~Koyejo}, \bibinfo{person}{S.~Mohamed}, \bibinfo{person}{A.~Agarwal}, \bibinfo{person}{D.~Belgrave}, \bibinfo{person}{K.~Cho}, {and} \bibinfo{person}{A.~Oh}} (Eds.), Vol.~\bibinfo{volume}{35}. \bibinfo{publisher}{Curran Associates, Inc.}, \bibinfo{pages}{23716--23736}.
\newblock


\bibitem[Batool et~al\mbox{.}(2025)]%
        {batool2025impedancegpt}
\bibfield{author}{\bibinfo{person}{Fatima Batool}, \bibinfo{person}{Muhammad Zafar}, \bibinfo{person}{Yasheerah Yaqoot}, \bibinfo{person}{R.~A. Khan}, \bibinfo{person}{M.~H. Khan}, \bibinfo{person}{Alexey Fedoseev}, {and} \bibinfo{person}{Dzmitry Tsetserukou}.} \bibinfo{year}{2025}\natexlab{}.
\newblock \showarticletitle{ImpedanceGPT: VLM-driven Impedance Control of Swarm of Mini-drones for Intelligent Navigation in Dynamic Environment}. In \bibinfo{booktitle}{\emph{Proc. of the IEEE/RSJ Int. Conf. on Intelligent Robots and Systems (IROS)}}. \bibinfo{address}{Hangzhou, China}, \bibinfo{pages}{2592--2597}.
\newblock


\bibitem[Bicchi and Kumar(2000)]%
        {bicchi2000roboticHands}
\bibfield{author}{\bibinfo{person}{A. Bicchi} {and} \bibinfo{person}{V. Kumar}.} \bibinfo{year}{2000}\natexlab{}.
\newblock \showarticletitle{Robotic grasping and contact: a review}. In \bibinfo{booktitle}{\emph{Proc. IEEE Int. Conf. on Robotics and Automation. Symposia Proceedings (Cat. No.00CH37065)}}, Vol.~\bibinfo{volume}{1}. \bibinfo{pages}{348--353 vol.1}.
\newblock
\href{https://doi.org/10.1109/ROBOT.2000.844081}{doi:\nolinkurl{10.1109/ROBOT.2000.844081}}


\bibitem[Brohan et~al\mbox{.}(2023)]%
        {brohan2023rt1roboticstransformerrealworld}
\bibfield{author}{\bibinfo{person}{Anthony Brohan}, \bibinfo{person}{Noah Brown}, \bibinfo{person}{Justice Carbajal}, \bibinfo{person}{Yevgen Chebotar}, \bibinfo{person}{Joseph Dabis}, \bibinfo{person}{Chelsea Finn}, {et~al\mbox{.}}} \bibinfo{year}{2023}\natexlab{}.
\newblock \bibinfo{booktitle}{\emph{{RT-1}: Robotics Transformer for Real-World Control at Scale}}.
\newblock
\showeprint[arxiv]{2212.06817}
\newblock
\shownote{Retrieved from https://arxiv.org/abs/2212.06817}.


\bibitem[Gao et~al\mbox{.}(2024)]%
        {gao2024_pgvlm}
\bibfield{author}{\bibinfo{person}{Jensen Gao}, \bibinfo{person}{Bidipta Sarkar}, \bibinfo{person}{Fei Xia}, \bibinfo{person}{Ted Xiao}, \bibinfo{person}{Jiajun Wu}, \bibinfo{person}{Brian Ichter}, \bibinfo{person}{Anirudha Majumdar}, {and} \bibinfo{person}{Dorsa Sadigh}.} \bibinfo{year}{2024}\natexlab{}.
\newblock \showarticletitle{Physically Grounded Vision--Language Models for Robotic Manipulation}.
\newblock \bibinfo{journal}{\emph{Proc. of the IEEE Int. Conf. on Robotics and Automation (ICRA)}}  \bibinfo{volume}{2024} (\bibinfo{year}{2024}), \bibinfo{pages}{12462--12469}.
\newblock
\href{https://doi.org/10.1109/ICRA57147.2024.10610090}{doi:\nolinkurl{10.1109/ICRA57147.2024.10610090}}


\bibitem[Gbagbe et~al\mbox{.}(2024)]%
        {gbagbe2024bivla}
\bibfield{author}{\bibinfo{person}{K.~Fidele Gbagbe}, \bibinfo{person}{Miguel Altamirano~Cabrera}, \bibinfo{person}{Ahmed Alabbas}, \bibinfo{person}{Omar Alyounes}, \bibinfo{person}{Alexey Lykov}, {and} \bibinfo{person}{Dzmitry Tsetserukou}.} \bibinfo{year}{2024}\natexlab{}.
\newblock \showarticletitle{Bi-VLA: Vision-Language-Action Model-Based System for Bimanual Robotic Dexterous Manipulations}. In \bibinfo{booktitle}{\emph{Proc. of the IEEE Int. Conf. on Systems, Man, and Cybernetics (SMC)}}. \bibinfo{address}{Sarawak, Malaysia}, \bibinfo{pages}{2864--2869}.
\newblock


\bibitem[Hogan(1985)]%
        {hogan1985_impedance_i}
\bibfield{author}{\bibinfo{person}{Neville Hogan}.} \bibinfo{year}{1985}\natexlab{}.
\newblock \showarticletitle{Impedance Control: An Approach to Manipulation: Part I—Theory}.
\newblock \bibinfo{journal}{\emph{Journal of Dynamic Systems, Measurement, and Control}} \bibinfo{volume}{107}, \bibinfo{number}{1} (\bibinfo{year}{1985}), \bibinfo{pages}{1--7}.
\newblock
\href{https://doi.org/10.1115/1.3140702}{doi:\nolinkurl{10.1115/1.3140702}}


\bibitem[Jeong et~al\mbox{.}(2024)]%
        {jeong2024_llm_robot_intel}
\bibfield{author}{\bibinfo{person}{Hyeongyo Jeong}, \bibinfo{person}{Haechan Lee}, \bibinfo{person}{Changwon Kim}, {and} \bibinfo{person}{Sungtae Shin}.} \bibinfo{year}{2024}\natexlab{}.
\newblock \showarticletitle{A Survey of Robot Intelligence with Large Language Models}.
\newblock \bibinfo{journal}{\emph{Applied Sciences}} \bibinfo{volume}{14}, \bibinfo{number}{19} (\bibinfo{year}{2024}), \bibinfo{pages}{8868}.
\newblock
\href{https://doi.org/10.3390/app14198868}{doi:\nolinkurl{10.3390/app14198868}}


\bibitem[Ji et~al\mbox{.}(2024)]%
        {ji2024_fm_hrc}
\bibfield{author}{\bibinfo{person}{Yuchen Ji}, \bibinfo{person}{Zequn Zhang}, \bibinfo{person}{Dunbing Tang}, \bibinfo{person}{Yi Zheng}, \bibinfo{person}{Changchun Liu}, \bibinfo{person}{Zhen Zhao}, {and} \bibinfo{person}{Xinghui Li}.} \bibinfo{year}{2024}\natexlab{}.
\newblock \showarticletitle{Foundation Models Assist in Human--Robot Collaboration Assembly}.
\newblock \bibinfo{journal}{\emph{Scientific Reports}}  \bibinfo{volume}{14} (\bibinfo{year}{2024}), \bibinfo{pages}{24828}.
\newblock
\href{https://doi.org/10.1038/s41598-024-75715-4}{doi:\nolinkurl{10.1038/s41598-024-75715-4}}


\bibitem[Li et~al\mbox{.}(2025)]%
        {li2025_fm_manip_survey}
\bibfield{author}{\bibinfo{person}{Dingzhe Li}, \bibinfo{person}{Yixiang Jin}, \bibinfo{person}{Yuhao Sun}, \bibinfo{person}{Yong A}, \bibinfo{person}{Hongze Yu}, \bibinfo{person}{Jun Shi}, \bibinfo{person}{Xiaoshuai Hao}, \bibinfo{person}{Peng Hao}, \bibinfo{person}{Huaping Liu}, \bibinfo{person}{Fuchun Sun}, \bibinfo{person}{Jianwei Zhang}, {and} \bibinfo{person}{Bin Fang}.} \bibinfo{year}{2025}\natexlab{}.
\newblock \showarticletitle{What Foundation Models Can Bring for Robot Learning in Manipulation: A Survey}.
\newblock \bibinfo{journal}{\emph{The International Journal of Robotics Research}}  \bibinfo{volume}{36} (\bibinfo{year}{2025}), \bibinfo{pages}{261--268}.
\newblock
\href{https://doi.org/10.1177/02783649251390579}{doi:\nolinkurl{10.1177/02783649251390579}}


\bibitem[Tsetserukou et~al\mbox{.}(2008)]%
        {tsetserukou2008tactile}
\bibfield{author}{\bibinfo{person}{Dzmitry Tsetserukou}, \bibinfo{person}{Naoki Kawakami}, {and} \bibinfo{person}{Susumu Tachi}.} \bibinfo{year}{2008}\natexlab{}.
\newblock \showarticletitle{Obstacle avoidance control of humanoid robot arm through tactile interaction}. In \bibinfo{booktitle}{\emph{Proc. of the IEEE-RAS Int. Conf. on Humanoid Robots}}. \bibinfo{address}{Daejeon, Korea}, \bibinfo{pages}{379--384}.
\newblock


\bibitem[Zafar et~al\mbox{.}(2025)]%
        {zafar2025swarmvlm}
\bibfield{author}{\bibinfo{person}{Muhammad Zafar}, \bibinfo{person}{R.~A. Khan}, \bibinfo{person}{Fatima Batool}, \bibinfo{person}{Yasheerah Yaqoot}, \bibinfo{person}{Zhi Guo}, \bibinfo{person}{Mikhail Litvinov}, \bibinfo{person}{Alexey Fedoseev}, {and} \bibinfo{person}{Dzmitry Tsetserukou}.} \bibinfo{year}{2025}\natexlab{}.
\newblock \showarticletitle{SwarmVLM: VLM-Guided Impedance Control for Autonomous Navigation of Heterogeneous Robots in Dynamic Warehousing}. In \bibinfo{booktitle}{\emph{Proc. of the IEEE Int. Conf. on Robotics and Biomimetics (ROBIO)}}. \bibinfo{address}{Chengdu, China}, \bibinfo{pages}{770--775}.
\newblock


\bibitem[Zhang et~al\mbox{.}(2025)]%
        {omnivic2025_vlm_vic}
\bibfield{author}{\bibinfo{person}{Heng Zhang}, \bibinfo{person}{Wei-Hsing Huang}, \bibinfo{person}{Gokhan Solak}, {and} \bibinfo{person}{Arash Ajoudani}.} \bibinfo{year}{2025}\natexlab{}.
\newblock \bibinfo{booktitle}{\emph{OmniVIC: A Self-Improving Variable Impedance Controller with Vision–Language In-Context Learning for Safe Robotic Manipulation}}.
\newblock
\href{https://doi.org/10.48550/arXiv.2510.17150}{doi:\nolinkurl{10.48550/arXiv.2510.17150}}
\newblock
\shownote{Retrieved from https://arxiv.org/abs/2510.17150}.


\bibitem[Zitkovich et~al\mbox{.}(2023)]%
        {brohan2023_rt2}
\bibfield{author}{\bibinfo{person}{Brianna Zitkovich}, \bibinfo{person}{Tianhe Yu}, \bibinfo{person}{Sichun Xu}, \bibinfo{person}{Peng Xu}, \bibinfo{person}{Ted Xiao}, \bibinfo{person}{Fei Xia}, \bibinfo{person}{Jialin Wu}, \bibinfo{person}{Paul Wohlhart}, \bibinfo{person}{Stefan Welker}, \bibinfo{person}{Ayzaan Wahid}, {et~al\mbox{.}}} \bibinfo{year}{2023}\natexlab{}.
\newblock \showarticletitle{RT-2: Vision--Language--Action Models Transfer Web Knowledge to Robotic Control}.
\newblock \bibinfo{journal}{\emph{Proc. of the 7th Conf.on Robot Learning}}  \bibinfo{volume}{229} (\bibinfo{year}{2023}), \bibinfo{pages}{2165--2183}.
\newblock


\end{thebibliography}


\end{document}